\pdfoutput=1

\documentclass[11pt]{article}

\usepackage{ACL2023}

\usepackage{times}
\usepackage{latexsym}

\usepackage[T1]{fontenc}

\usepackage[utf8]{inputenc}

\usepackage{microtype}

\usepackage{inconsolata}

\usepackage{graphicx}
\usepackage{xcolor}
\usepackage{microtype}
\usepackage{enumitem}
\usepackage{algorithm}
\usepackage{algorithmic}
\usepackage{amsmath}
\usepackage{multirow}
\usepackage{booktabs}
\usepackage{url}
\usepackage{graphicx}
\definecolor{RoseQuartzBg}{HTML}{F7CAC9}
\definecolor{RoseQuartz}{HTML}{F5A798}
\definecolor{Serenity}{HTML}{92A8D1}
\definecolor{OrangeRed}{rgb}{1.0, 0.27, 0.0}
\definecolor{Red}{rgb}{1.0, 0.0, 0.0}
\definecolor{Turquoise}{HTML}{0F4C81}
\usepackage{xparse}
\usepackage{soul}
\usepackage{xspace}
\NewDocumentCommand{\alex}{ mO{} }{\textcolor{OrangeRed}{\textsuperscript{\textit{Alex}}\textsf{\textbf{\small[#1]}}}}
\NewDocumentCommand{\patrick}{ mO{} }{\textcolor{purple}{\textsuperscript{\textit{Patrick}}\textsf{\textbf{\small[#1]}}}}
\NewDocumentCommand{\sijia}{ mO{} }{\textcolor{blue}{\textsuperscript{\textit{Sijia}}\textsf{\textbf{\small[#1]}}}}
\NewDocumentCommand{\pramudi}{ mO{} }{\textcolor{red}{\textsuperscript{\textit{Pramuditha}}\textsf{\textbf{\small[#1]}}}}
\NewDocumentCommand{\sheng}{ mO{} }{\textcolor{green}{\textsuperscript{\textit{Sheng}}\textsf{\textbf{\small[#1]}}}}
\NewDocumentCommand{\zhiguow}{ mO{} }{\textcolor{orange}{\textsuperscript{\textit{Zhiguo}}\textsf{\textbf{\small[#1]}}}}
\NewDocumentCommand{\chungwei}{ mO{} }{\textcolor{brown}{\textsuperscript{\textit{Chung-Wei}}\textsf{\textbf{\small[#1]}}}}
\NewDocumentCommand{\ww}{ mO{} }{\textcolor{teal}{\textsuperscript{\textit{William}}\textsf{\textbf{\small[#1]}}}}
\NewDocumentCommand{\henry}{ mO{} }{\textcolor{teal}{\textsuperscript{\textit{Henry}}\textsf{\textbf{\small[#1]}}}}
\NewDocumentCommand{\jiarong}{ mO{} }{\textcolor{red}{\textsuperscript{\textit{Jiarong}}\textsf{\textbf{\small[#1]}}}}

\usepackage{amssymb}
\usepackage{array}
\newcolumntype{L}{>{\centering\arraybackslash}m{4.5cm}}
\newcolumntype{V}{>{\centering\arraybackslash}m{6.5cm}}
\newcolumntype{M}{>{\centering\arraybackslash}m{3cm}}
\newcolumntype{A}{>{\centering\arraybackslash}m{4cm}}
\newcolumntype{N}{>{\arraybackslash}m{3cm}}
\newcolumntype{S}{>{\centering\arraybackslash}m{2cm}}
\newcolumntype{X}{>{\arraybackslash}m{2cm}}
\newcolumntype{P}{>{\arraybackslash}m{10cm}}
\newcolumntype{Q}{>{\arraybackslash}m{5cm}}
\newcommand\RaiseImage[2][scale=1]{%
  \raisebox{-0.5\totalheight}{\includegraphics[#1]{#2}}}
  
\newcommand{\Modelsp}{\textsc{GDMM} }
\newcommand{\Model}{\textsc{GDMM}\xspace}

\newcommand{\task}{\textsc{DMEL}\xspace}
\newcommand{\tasksp}{\textsc{DMEL} }
\usepackage{array}
\usepackage{makecell}

\title{Benchmarking Diverse-Modal Entity Linking with Generative Models}

\author{Sijia Wang\textsuperscript{\rm 1}\thanks{\ \ Work conducted during an internship at Amazon.} \hspace{0.1em},
        Alexander Hanbo Li\textsuperscript{\rm 2}\thanks{ \ \ Corresponding author.}, 
        Henry Zhu\textsuperscript{\rm 2}, 
        Sheng Zhang\textsuperscript{\rm 2}, 
        Chung-Wei Hang\textsuperscript{\rm 2}, \\
        {\bf Pramuditha Perera\textsuperscript{\rm 2}, 
        Jie Ma\textsuperscript{\rm 2},
        William Wang\textsuperscript{\rm 2}, 
        Zhiguo Wang\textsuperscript{\rm 2}, 
        Vittorio Castelli\textsuperscript{\rm 2}} \\
        {\bf Bing Xiang\textsuperscript{\rm 2}, 
        Patrick Ng\textsuperscript{\rm 2}} \\
        \textsuperscript{\rm 1} Virginia Tech \quad  \textsuperscript{\rm 2} AWS AI Labs\\
        \texttt{\footnotesize{sijiawang@vt.edu}} \\
        \texttt{\footnotesize{\{hanboli,henghui,zshe,cwhang,pramudi,jieman,wyw,
        zhiguow,vittorca,bxiang,patricng\}@amazon.com}}
        }

\date{}

\begin{document}
\maketitle
\begin{abstract}
Entities can be expressed in diverse formats, such as texts, images, or column names and cell values in tables. {While existing entity linking (\textbf{EL}) models work well on per modality configuration, such as text-only EL, visual grounding, or schema linking},
it is more challenging to design a unified model for diverse modality configurations.
To bring various modality configurations together, we constructed a benchmark for diverse-modal EL (\textbf{\task}) from existing EL datasets, covering all three modalities including text, image, and table. To approach the \task task, we proposed a generative diverse-modal model (\textbf{\Model}) following a multimodal-encoder-decoder paradigm. Pre-training \Model with rich corpora builds a solid foundation for \task without storing the entire KB for inference. Fine-tuning \Model builds a stronger \task baseline, outperforming state-of-the-art task-specific EL models by 8.51 F1 score on average.
Additionally, extensive error analyses are conducted to highlight the challenges of \task, facilitating future research on this task.

\end{abstract}

\section{Introduction}
\label{sec:intro}

\begin{figure}[t]
\includegraphics[width=7.6cm]{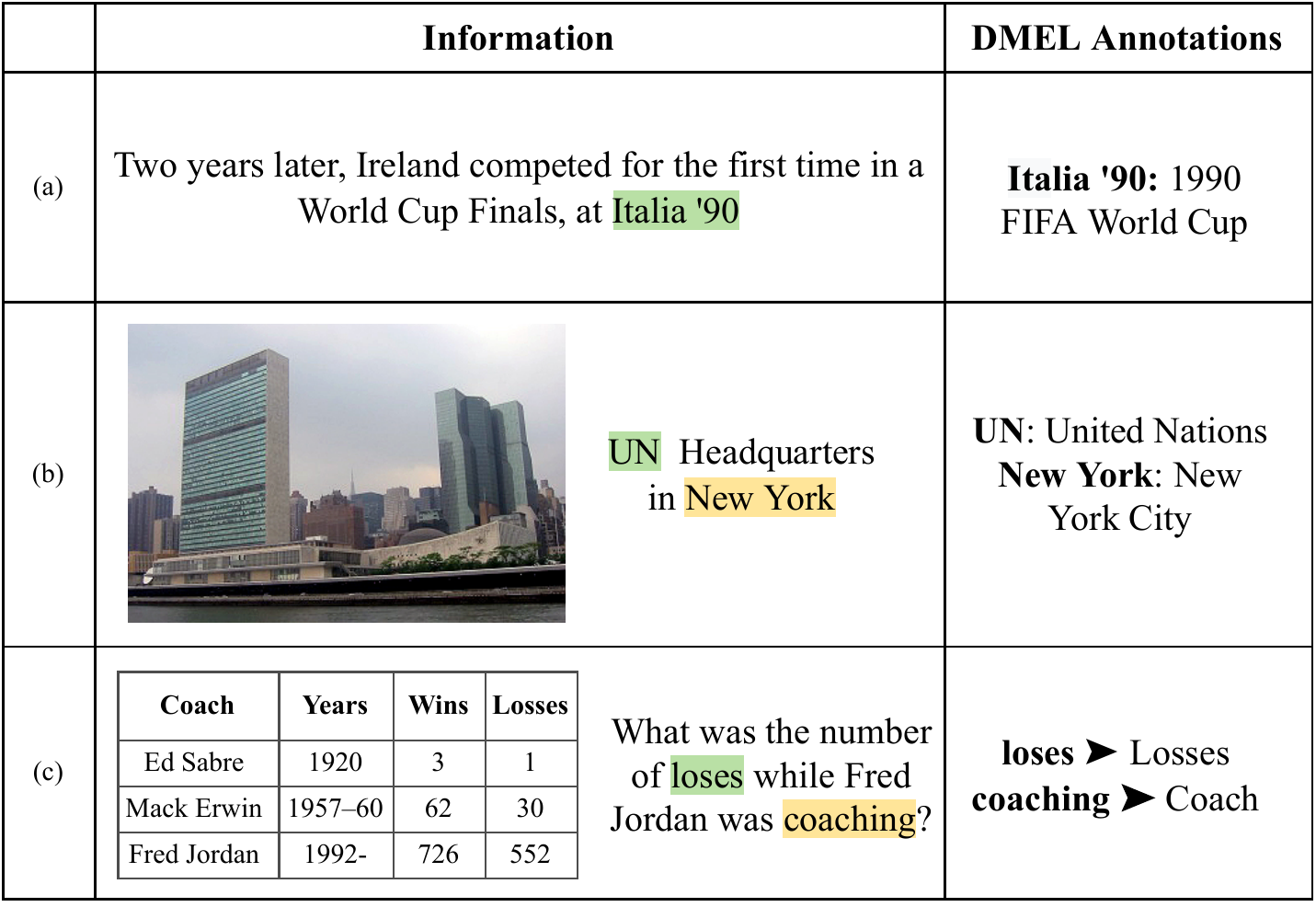}
\centering
\caption{DMEL examples for (a) textual EL, (b) textual-visual EL, and (c) tabular schema linking.}
\label{fig:intro_example}
\end{figure}

Linking ambiguous mentions to unambiguous referent in a knowledge base (KB) such as Wikipedia, known as \textbf{Entity linking} (EL) \cite{6823700}, is an essential component for applications like question answering \cite{6177724, chen-etal-2017-reading, NEURIPS2020_6b493230} and recommendation systems \cite{yang-etal-2018-collective}.
\textbf{Diverse-Modal Entity Linking} (\task) extends the scope of interest from textual entity linking to heterogeneous input formats, such as linking visual and textual expressions to KB \cite{twitter_mel1, twitter_mel2, moon-etal-2018-multimodal-snap, gan_2021_melbench, Zheng_2022_veld,wang-etal-2022-wikidiverse, gan_2021_melbench, cui2021:whos-waldo} and linking mentions in natural language to tables or database (DB) schema \cite{liu-etal-2021-awakening, Katsakioris_2022_tabular_el_needs_attention, shi-etal-2020-squall, lei-etal-2020-slsql, chen-etal-2020-tale, Wang2022ACM_proton}. 
{Figure~\ref{fig:intro_example} demonstrate three examples of {\task, including} (a) classical textual entity linking, (b) textual-visual entity linking in which the question or mentions are paired with image(s), and (c) tabular schema linking in which the mentions are linked to column names or cell values.}

{Retrieval-based} contrastive learning or ranking mechanism is the mainstream for early visual entity linking by leveraging a matching score between the mention and the KB entities \cite{cui2021:whos-waldo,wang-etal-2022-wikidiverse, Zheng_2022_veld}. 
{However, these methods require storage of dense representations of all KB entities, and when the size of entities increases (e.g. Wikipedia has ~6M articles), it raises concerns for space complexity and also inference-time latency.}
Meanwhile, linking mentions to tables or DB schemes, known as schema linking, remains an important but under-explored task. {For example, in text-to-SQL generations, incorrect schema linking usually counts for a large portion of the errors \citep{zhongSeq2SQL2017,yu-etal-2018-spider,shi-etal-2020-squall,lei-etal-2020-slsql,Taniguchi2021AnIB}.} Previous string matching heuristic \cite{chen-etal-2020-tale} or embedding matching methods \cite{chen-etal-2020-tale, Wang2022ACM_proton, Guo2019TowardsCT, wang-etal-2020-ratsql} lack semantic and schema understanding, and can hardly generalize well to new domains. 
Last but not the least, previous endeavors of entity linking are limited to individual tasks including textual EL, textual-visual EL, or schema linking, and lack a general view for the \tasksp problem.

{To this end, we propose a unified DMEL task that includes existing EL datasets on all three modalities -- text, image, and table.}
The unified \tasksp task is challenging because the model needs to handle a wide spectrum of modality configurations together. 
On the modeling side, {because storing all entity information (e.g. all the images in the entire KB) is expensive at inference time, we propose to use a unified generative model that can take diverse-modal input and generate entity names in an autoregressive fashion.} 
Additionally, the mention diversity and ambiguity issue in schema linking can be addressed by pre-training the generative model. 

In this work, we build a generic diverse-modal 
architecture for end-to-end \task.  The \tasksp dataset is constructed from five existing datasets, including GERBIL benchmark, WikiDiverse, MELBench-Wikipedia, Squall, and SLSQL, covering diverse EL tasks. 
The proposed generative diverse-modal model (\textbf{\Model}) is first pre-trained on large-scale text corpus BLINK and images corpus from Wikipedia KB, offering profound prior knowledge.
Extensive experiments are then conducted on the \tasksp benchmark to compare our proposed generative model to previous state-of-the-art methods.  Experimental results show that \Modelsp achieves strong performance on the \tasksp dataset and outperforms state-of-the-art task-specific EL models. Our contributions include: 
\begin{itemize} 
    \item We define a novel diverse-modal Entity Linking task, which links an entity mention within heterogeneous information sources to a knowledge base. 
          A unified dataset  is constructed for rigorous \tasksp examination. 
    \item {A generative diverse-modal model \Modelsp is proposed following a multimodal-encoder-decoder structure. The multimodal encoder allows collective representation between each modality. The autoregressive structure enables us to directly predict the entity name without storing the entire KB. The pre-training experimental results confirm that a candidate trie created from entity names is sufficient for inference.}
    \item The experimental results show that the proposed model obtains state-of-the-art performance on (almost) each individual EL task.
\end{itemize}


\section{Problem Formulation} 
\label{sec:problem}
We assume to have a KB (e.g., Wikipedia or a DB schema) where each entity is a unique entry in the KB. We formulate the following \task problem:
given a multimodal input $\{\mathbf{x}_i, \mathbf{v}_i, \mathbf{u}_i\}$ of textual (L), visual (V), and tabular (U) modality respectively, an entity mention $\mathbf{m}_i$ within the input, and a candidate set $\mathcal{C}_i = \{\mathbf{c}_i^1, \cdots, \mathbf{c}_i^K\}$, the task is to link the mention $\mathbf{m}_i$ to one entity in $\mathcal{C}_i$. {We assume the entity span is given. Sometimes the candidate set can be the entire entity collection $\mathcal E$.}
{Particular instances of \task problem include but are not limited to: 
\textit{Textual Entity Disambiguation} where a given mention $\mathbf{m}_i$  in $\mathbf{x}_i$ will be linked to one entity in $\mathcal{C}_i$;  
\textit{Textual-Visual Entity Disambiguation} where the $\mathbf{m}_i$ and a given image $\mathbf{v}_i$ will be linked to one entity in  $\mathcal{C}_i$;  
\textit{Schema linking} where a $\mathbf{m}_i$ in a SQL query will be linked to table schema, i.e., a column name within given tables $\mathbf{u}_i$.}
{If the mention is not a valid entity or not in $\mathcal{C}_i$, the target label is ``nil''.}

\begin{table}[t]
\centering
\small
\scalebox{1.0}{
\begin{tabular}{lcccc}
\toprule
    \multirow{2}{*}{\textbf{Dataset}} & \multirow{2}{*}{\textbf{Modality}}
    & \multicolumn{3}{c}{ \textbf{Size}}\\
\cmidrule(lr){3-5}
    && \textbf{\#L} & \textbf{\#V} & \textbf{\#U}  \\
\midrule
    GERBIL      & L → L &  42,854         &  0     &0      \\
    WikiDiverse	& LV → L&    7,823      &  6,924     & 0     \\	
    MELBench    & LV → L&	18,880      &  18,880 & 0     \\ 
    Squall	    & LU → L&	11,274      &  0    & 2,108      \\
    SLSQL	    & LU → L&	8,034       &  0    & 166      \\ 
\midrule
    \textbf{\task} & LVU → L& 88,865  & 25,804 & 2,274\\
\bottomrule
\end{tabular}}
\caption{Comparison of existing datasets. Statistics for modality, (L)anguage, (V)ision, and Tabl(U)r, are shown. 
}
\label{tab:data_modal}
\end{table}

\begin{figure*}[t]
\includegraphics[width=15cm]{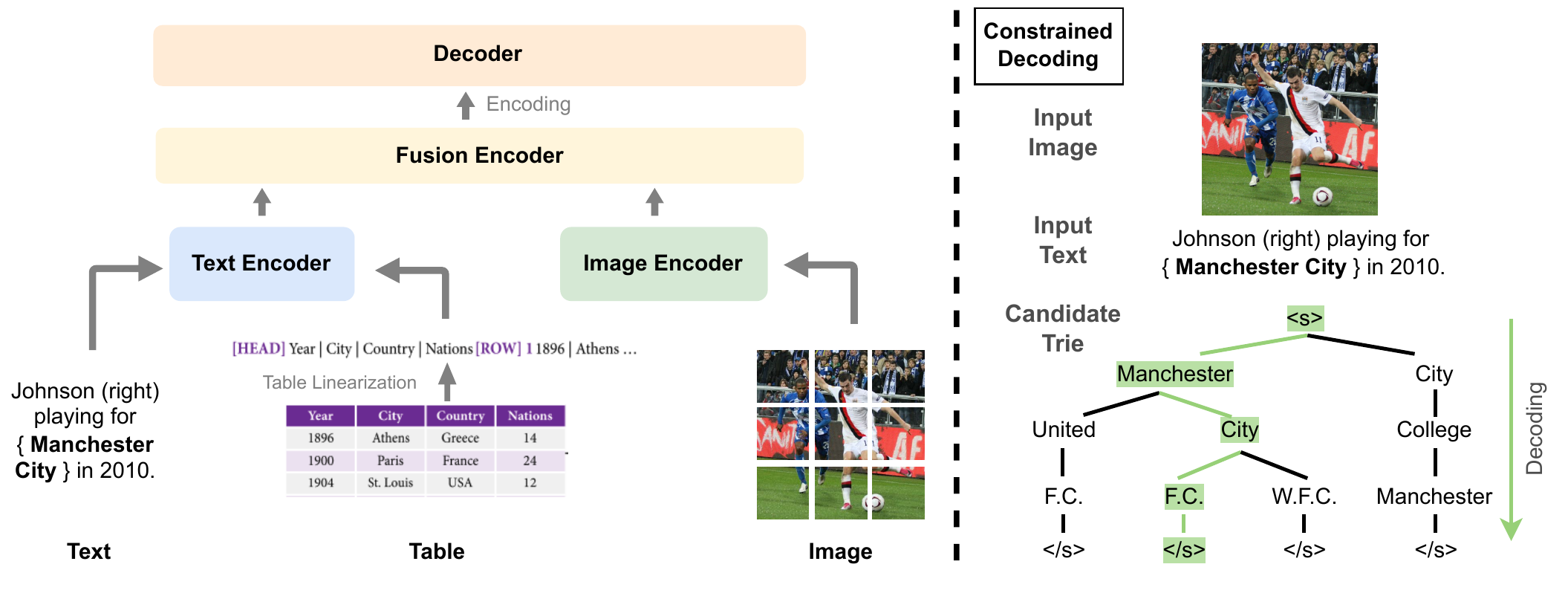}
\centering
\caption{Overall \Modelsp architecture (left) and constrained decoding illustration (right).}
\label{fig:model}
\end{figure*}

\section{\tasksp Benchmark}

{We build the \tasksp benchmark from five existing datasets, including} GERBIL benchmark~\cite{GERBIL2018}, WikiDiverse~\cite{wang-etal-2022-wikidiverse},  MELBench-Wikidata (as MELBench in the rest of the paper)~\cite{gan_2021_melbench}, Squall~\cite{shi-etal-2020-squall}, and SLSQL~\cite{lei-etal-2020-slsql}. 
{We evaluate textual-visual entity disambiguation capability on WikiDiverse and MELBench, and evaluate tabular schema linking on Squall and SLSQL. All datasets are in English, and we summarize the data statistics in Table \ref{tab:data_modal}. {We will release this benchmark.}}

We assume that the mention span is given across all datasets, thus   
(1) on GERBIL benchmark, we investigate textual entity disambiguation using the same candidate sets as in \citet{le-titov-2018-improving} and \citet{decao2021genre}. 
(2) On WikiDiverse, we investigate entity disambiguation performance with retrieved Top-10 candidates by \citet{wang-etal-2022-wikidiverse}
(3) On MELBench-Wikidata, we follow the original setting in which the given entity mention will be linked to its referent in the knowledge base. 
(4) On Squall, the given entity mention in the natural question will be linked to column names in the target table. 
(5) On SLSQL,
the entity mention in the natural question will be linked to the column name within multiple tables.


Statistics for each individual EL task are shown in Table \ref{tab:data}. Note that for the GERBIL benchmark, the training split refers to AIDA-train, the validation split refers to AIDA-dev, and the test split includes all test split as in \cite{decao2021genre}.

\begin{table}[t]
\centering
\small{
\begin{tabular}{l|c|rrr}
\toprule
    \multirow{2}{*}{\textbf{Dataset}} & \multirow{2}{*}{\textbf{Modality}}
    & \multicolumn{3}{c}{ \textbf{\# Sentences / Tables}}\\
    && \textbf{Train} & \textbf{Valid} &\textbf{Test} \\
\midrule
    GERBIL & L → L &18,448&4,791 & 19,614\\
    WikiDiverse	& VL → L    & 6,312   & 755	  &757\\	
    MELBench    & VL → L	& 13,216 & 1,888  & 3,776\\ 
    Squall	    & LU → L	& 9,028  & 2,246  & - \\
    SLSQL	    & LU → L	& 7,000  & 1,034  & -\\ 
\bottomrule
\end{tabular}}
\caption{Data statistics. Modalities include (L)anguage, (V)ision, and Tabl(U)r modality. 
}
\label{tab:data}
\end{table}

\section{GDMM Model}
\label{sec:model}

We build a generative entity linking model that enables diverse-modal 
vision, language and table understanding and inference. We show how this can be achieved with a generative encoder-decoder structure. 


\subsection{Input Processor}
\label{sec:input_processor}
As shown in Figure \ref{fig:model}, our model can process inputs of three modalities, including texts, images, and tables. Formally, given a multimodal input $\{\mathbf{x}_i, \mathbf{v}_i, \mathbf{u}_i\}$,
the input processor serves to encode the data and group the modalities as follows.
(1) \textbf{Text} Given an input text $\mathbf{x}_i$, we first tokenize and embed it into a list of word vectors $\pmb{x}_i$ following \citet{devlin-etal-2019-bert}.
(2) {\textbf{Image}} Given an input image $\mathbf{v}_i$, we first resize it to a fixed size and split it into patches, following \citet{vilt}.
(3) {\textbf{Table}}
We follow the table representation proposed in TAPEX \cite{liu2022tapex}. The table is flattened and represented as $\mathbf{u}_i^*=\texttt{[head]}, col_1, \allowbreak\cdots, col_M, \texttt{[ROW]}, 1,  cell_{11}, \cdots, cell_{1M}, \texttt{[ROW]} $ $\cdots$, 
where $\texttt{[head]}$ and $\texttt{[ROW]}$ are special tokens denoting the beginning of table headers and rows, and the number after $\texttt{[ROW]}$ is used to denote the row index. {$cell_{ij}$ represents a cell in $i$th row and $j$th column.} Then the table representation $\mathbf{u}_i^*$ will be tokenized and embedded into $\pmb{u}_i$.
Finally, given the multimodal input $\{\mathbf{x}_i, \mathbf{v}_i, \mathbf{u}_i\}$, our input processor outputs $\{\pmb{x}_i\bigoplus \pmb{u}_i, \pmb{v}_i\}$, where $\bigoplus$ is a concatenation operator.

\subsection{\Model Model Architecture}
\paragraph{Multimodal Encoder}
The multimodal encoder consists of an image encoder, a text encoder, and a fusion encoder, following previous work \cite{flava,Yang2022tcl}. The text encoder and image encoder use the same ViT architecture \cite{dosovitskiy2021vit} with different parameters. The text input and visual input $\{\pmb{x}_i\bigoplus \pmb{u}_i, \pmb{v}_i \}$ are passed into the text encoder and vision encoder individually. The text hidden state vectors  $\{\pmb h_i^{LU}\}$ and the image embeddings $\{\pmb h_i^V\}$ are then projected and concatenated into a single list. A fusion encoder is applied to the concatenated list, which allows cross-attention between the projected unimodal representations and fuses the two. The output is a list of hidden states $\{\pmb{h}^M_i\}$.
The multimodal encoder parameters are initialized with pre-trained FLAVA \cite{flava} parameters.

\paragraph{Decoder} 
We exploit the transformer architecture \cite{Vaswani2017attention} for the decoder. {Previous study \cite{rothe-etal-2020-leveraging} points out that combining models with same vocabulary has stronger overall performance, thus we} initialize the decoder with the BERT~\cite{devlin-etal-2019-bert} pre-trained parameters. 

\paragraph{Training and Inference} The \Model is trained with standard autoregressive objective, i.e., maximizing the output sequence likelihood $p_\theta (\pmb y_i|\pmb{x}_i, \pmb{v}_i, \pmb{u}_i)$ with respect to the model's parameters $\theta$.
We rank each candidate $\mathbf{c}_i^k\in \mathcal{C}_i$ 
by computing a score with an autoregressive formulation:
$
    \text{score} (\mathbf{c}_i^k|\pmb{x}_i, \pmb{v}_i, \pmb{u}_i) 
    =p_\theta (\pmb y_i^k|\pmb{x}_i, \pmb{v}_i, \pmb{u}_i) 
    =\prod_{j=1}^N p_\theta ( y_j^k|y_{<j}^k, \pmb{x}_i, \pmb{v}_i, \pmb{u}_i),
$ where $N$ is the number of tokens of $\mathbf{c}_i^k$.
{If the score is lower than a threshold $\theta$, the prediction becomes ``nil''. The threshold will be decided by the development set.}

\paragraph{Constrained decoding} 
{When the candidate set  $\mathcal{C}_i$  is very large (e.g., the entire entity space $\mathcal{E}$), naturally, it is intractable to compute a score for every element. Thus we exploit Constrained Beam Search~\cite{Sutskever2014beamsearch, decao2021genre}, a tractable decoding strategy to efficiently search the valid entity space. 
It is tractable as the average time cost depends on beam size and the average length of entity representations (e.g. 6 BPE tokens on average for entities in Wikipedia KB), instead of the size of $\mathcal{C}_i$. An entity trie $\mathcal{T}_i$ for $\mathcal{C}_i$ will be created so that the output is limited to the target space. }
The constraint is defined as, for each node $t\in\mathcal{T}$, its children indicate all allowed continuations from the prefix traversing from root to $t$. For example, as shown in Figure~\ref{fig:model}, given four candidates \texttt{Manchester United F.C.}, \texttt{Manchester City F.C.}, \texttt{Manchester City W.F.C}, and \texttt{City College Manchester}, a candidate trie will be created as shown in the figure. The decoding will strictly follow the top-down order in the trie with a certain beam size.

\subsection{Pre-training \Model}

Pre-training is critical to our architecture though the encoder and decoder are initialized with pre-trained weights because the mapping between the encoder and decoder are randomly initialized and they have not been pre-trained simultaneously with the encoders and the decoder.

\paragraph{Pre-training data}
A pre-training corpus is constructed from BLINK \cite{Wu2020blink} and images in Wikipedia KB.
BLINK is a commonly used corpus for textual entity linking pre-training, including 9M unique annotations of document-mention-entity triples from Wikipedia. Meanwhile, the images in Wikipedia KB are naturally linked to their respective entity names. The two together are well-suited for pre-training \task models.
Aside from \textbf{text-only} BLINK, we construct \textbf{LV-paired} pre-training data by linking BLINK and Wiki-images. An image pool (Wiki-images)  is collected from Wikipedia KB if the entity can be linked to mentions in BLINK. The image pool contains 797,436 downloaded images of 495,149 entities in Wikipedia KB. We then randomly attach an image of the target entity, if exists, to each mention in BLINK. In total, the LV-paired pre-training data includes 5,445,264 mentions and 678,385 distinct images in the training set, and 5,816 mentions and 5,414 images in the development set. 


\paragraph{Pre-training details}

{
We pre-train \Model on text-only BLINK and LV-paired pre-training data in two stages. Note that not all the BLINK entities appear in Wiki-images. There are over 2.5M BLINK mentions not covered by LV-paired pre-training data. To fully leverage the BLINK annotations, we first pre-train on text-only BLINK and then pre-train on the LV-paired data. With text-only BLINK, we freeze parameters in the image encoder and fusion layers and only update parameters in the text encoder and decoder. In the second stage, all the parameters are updated.
}

\subsection{{Unified Learning}}
\label{sec:unified learning}
Upon the pre-trained model, one straightforward strategy for downstream tasks is single-task fine-tuning. We take one step further and investigate unified learning. Specifically, we'll investigate (1) single-task finetuning (ST-F), which refers to finetuning on individual tasks; (2) multi-task fine-tuning (MT-F) which combines the mixed training data of all datasets \cite{Raffel_t5}; and (3) multitask fine-tuning with prefixes (MT-FP), where we prepend task-specific prefixes like ``entity linking'' and ``schema linking'' to the input context.

\begin{table*}[t]
\centering
\small
\scalebox{0.9}{

\begin{tabular}{c|c|c|c|cc}
\toprule
    \textbf{Data} & \textbf{Task} & \textbf{Modality} & \textbf{Previous SOTA} & \textbf{\Model-base}&\textbf{\Model-large}\\
\midrule
    GERBIL & ED & L → L & \textbf{88.8}~\cite{decao2021genre} & 86.11{\scriptsize$\pm$0.24} &82.57{\scriptsize$\pm$0.22} \\
    WikiDiverse & VED & LV → L & 71.07~\cite{wang-etal-2022-wikidiverse} & \textbf{79.10}{\scriptsize$\pm$0.35}& 78.69{\scriptsize$\pm$0.33} \\
    MELBench &VED& LV → L &40.5~\cite{gan_2021_melbench} & 68.01{\scriptsize$\pm$0.75}& \textbf{72.41}{\scriptsize$\pm$0.65}\\
    Squall & SL& LU → L & 82.10{\scriptsize$\pm$2.41} (GENRE+)& \textbf{89.69}{\scriptsize$\pm$0.77}&89.12{\scriptsize$\pm$1.03}\\
    SLSQL & SL & LU → L & 82.80 (GENRE+)&81.48{\scriptsize$\pm$1.06} &\textbf{84.43}{\scriptsize$\pm$0.92}\\
\midrule
\multicolumn{3}{c|}{Avg.}& 72.93&80.88& 	81.44\\
\bottomrule
\end{tabular}}
\caption{{Benchmark results on \task data}
}
\label{tab:benchmark}
\end{table*}
\section{Experiments}
\label{sec:experiments}



\paragraph{Model Variants}
{We primarily report results on two model variants:
\textbf{\Model-base} where the decoder is initialized with BERT-base parameters, and 
\textbf{\Model-large} where the decoder is initialized with BERT-large parameters.
To investigate which modality provides dominant information for visual-text entity linking,  three configurations are explored:
\textbf{L+V} where both visual and textual information are given, \textbf{L} where only textual input are given, and \textbf{V} where only image are given. 
We report experimental results with a \textbf{single generic model} for the three modality settings. It is achieved by randomly masking out one modality during training. }
Implementation details are in Appendix \ref{sec:Implementation}.


\begin{figure}[t]
\includegraphics[width=7.5cm]{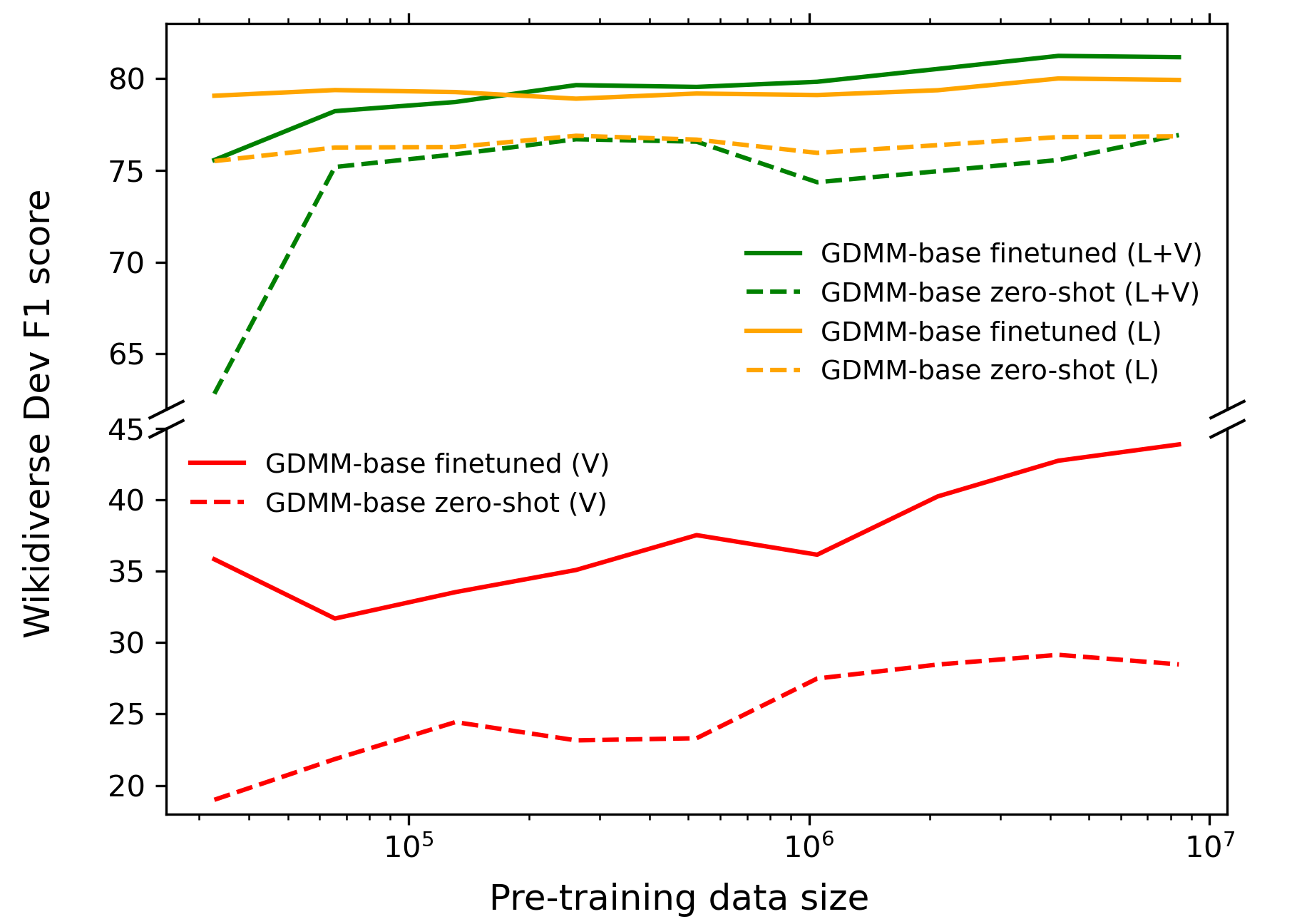}
\centering
\caption{The effect of the number of paired BLINK and Wiki-images pre-training data on \Model-base. 
}

\label{fig:pretrain}
\end{figure}

\subsection{Results}
\paragraph{Pre-training}

The pre-training of \Modelsp consists of two {stages}, text-only pre-training and LV-paired pre-training. {The pre-training performance is investigated with two methods: zero-shot or fine-tuned on WikiDiverse. Zero-shot refers to directly evaluating on WikiDiverse without training, while fine-tuned refers to further fine-tuning on the WikiDiverse.}
{The first-stage pre-trained checkpoint is directly evaluated on WikiDiverse with only text information (note that WikiDiverse is a dataset with both image and text input), achieving 75.43 zero-shot F1 score. {It greatly outperform the baseline model in \cite{wang-etal-2022-wikidiverse} by 4.36 F1 score, even though only text information are leveraged.} 
The evaluation  demonstrates that pre-training on BLINK builds a strong foundation for the proposed model.  
After that, we investigate the effect of paired pre-training data size in the second stage and visualize it in Figure \ref{fig:pretrain}.} It shows that the pre-training data size has a positive effect on inference with both image and text modality (L+V) and with only image modality (V). Text-only (L) performance is not affected even though the visual modality is introduced in the second pre-training stage.

\paragraph{Experimental results on \task benchmark} 
{
The experimental results of the proposed \Model on \task are shown in Table \ref{tab:benchmark}.
We compare  visual-language entity disambiguation result on WikiDiverse with LXMERT \cite{wang-etal-2022-wikidiverse}, and  visual-language entity linking performance on MELBench with \citet{gan_2021_melbench}. \Model achieves better performance on both datasets, especially on MELBench.
\Model strikingly improves the F1 score by over 31\%, demonstrating the effectiveness of the proposed architecture. 
}

{The schema linking performance is evaluated on Squall and SLSQL. For schema linking, we compare our model with the baseline model GENRE~\cite{decao2021genre}, as it has competitive performance in entity disambiguation. }
For a fair comparison, we fine-tune GENRE with their pre-trained checkpoint on BLINK and investigate two options, one without a flattened table (GENRE) and another with the flattened table (GENRE+) where the table content is leveraged identically as in \Model. Only experimental results for GENRE+ are reported in Table \ref{tab:benchmark} since it has better performance than GENRE.
The fact that the flattened table has better performance demonstrates the effectiveness of the table representation. Detailed results can be found in Appendix \ref{sec:appendix:experiments}.

\begin{table}[t]
\centering
\small
\scalebox{0.87}{
\begin{tabular}{l|cccc}
\toprule
    \multirow{2}{*}{\textbf{Dataset}} &  \multicolumn{4}{c}{\textbf{\Model-base}} \\
    & ZS & ST-F &MT-F &MT-FP \\
\midrule  
GERBIL      & 84.00     & \textbf{93.75} {\scriptsize$\pm$0.26}            &  \ul{93.63}  {\scriptsize$\pm$0.14}    & 93.56{\scriptsize$\pm$0.52}\\
WikiDiverse & 76.92     & \textbf{80.97}{\scriptsize$\pm$0.39}      & 80.02{\scriptsize$\pm$0.29}      & \ul{80.65}{\scriptsize$\pm$0.35} \\
MELBench    & 54.76     & \textbf{67.41}{\scriptsize$\pm$0.97}              &  63.64{\scriptsize$\pm$2.04}     & \ul{65.64}{\scriptsize$\pm$1.44} \\
SQUALL      & 47.52     & \textbf{89.69}{\scriptsize$\pm$0.77}     & 88.00{\scriptsize$\pm$1.31}       & \ul{88.37}{\scriptsize$\pm$0.99} \\
SLSQL       & 30.92     & 81.48{\scriptsize$\pm$1.06}               & \ul{83.59}{\scriptsize$\pm$1.90}     &\textbf{83.60}{\scriptsize$\pm$0.85} \\
\midrule
Avg.& 58.82  & 	\textbf{82.66} &81.78&\ul{82.36} \\
\bottomrule
\end{tabular}}
\caption{Unified learning results. 
Best scores and second-best scores are highlighted in \textbf{Bold} and \ul{underlined}. Variance does not apply to zero-shot F1 scores because the pre-trained checkpoint is unique.  We report results on the development set. 
}
\label{tab:multi}
\end{table}

\paragraph{Unified learning}
As mentioned in Section \ref{sec:unified learning}, we report unified learning results for ST-F, MT-F, and MT-FP in Table \ref{tab:multi}. To confirm whether the pre-trained checkpoints build a competitive foundation for visual-language entity linking, zero-shot (ZS) performance is also reported in the same table. 
The pre-trained checkpoint is competitive because the zero-shot performance (i.e. ZS column) outperforms previous state-of-the-art fine-tuned results for WikiDiverse in Table \ref{tab:wikidiverse} and MELBench in Table \ref{tab:melbench}. 
On average, ST-F and MT-FP achieve the best and the second-best performance, with a small gap between the two.
It is expected that ST-F achieves the best performance as each fine-tuned model is able to fit the target dataset distribution. Considering ST-F trains five models while MT-FP trains a single model, the competitive MT-FP performance suggests that model efficiency can be achieved at the cost of a minor performance drop, that is, 0.30  average F1 score drop for \Model-base.  
Additionally, the fact that MT-FP constantly outperforms MT-F aligns with previous findings that task-specific prefixes is effective in informing the model of the target tasks \cite{Dong2019UNILM, Raffel_t5}. 

\begin{figure}[t]
\includegraphics[width=6.5cm]{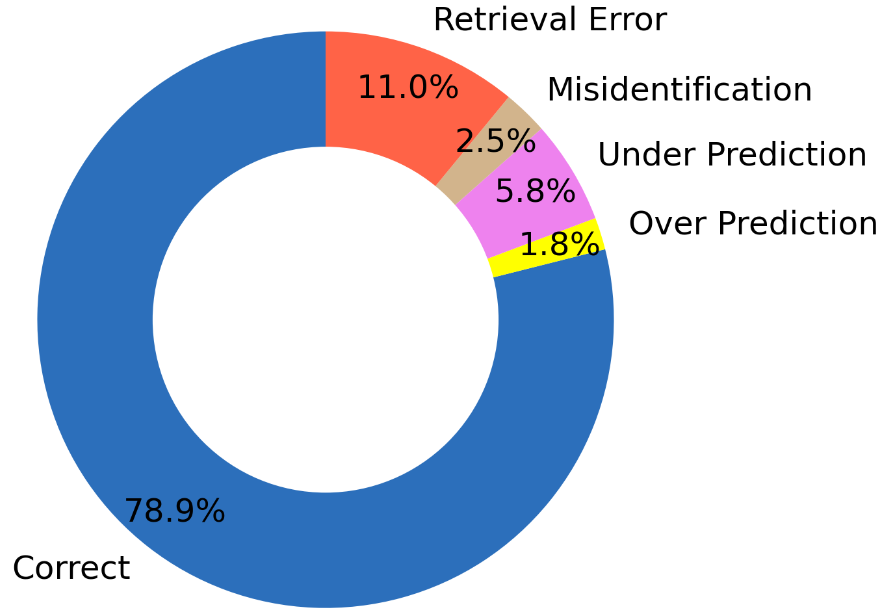}
\centering
\caption{Error breakdown on WikDiverse validation set.}
\label{fig:error_breakdown}

\end{figure}



\begin{table*}[ht!]
\begin{center}
\scalebox{0.7}{
\begin{tabular}{SLLAA}
\toprule
\textbf{ID} & (a) & (b) & (c) & (d)
    \\ 
\textbf{Image}
    &\RaiseImage[height=2.0cm]{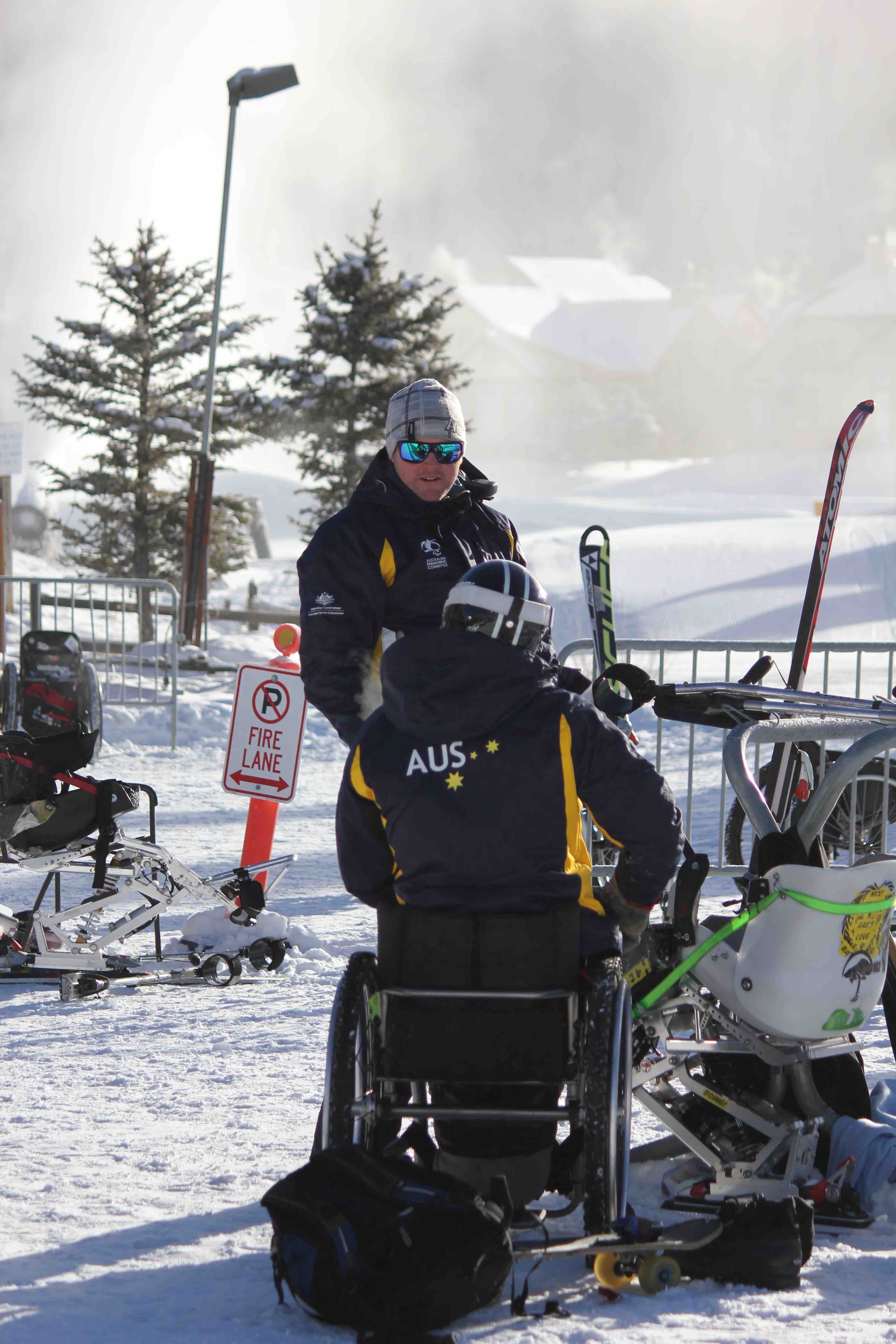}
    &\RaiseImage[height=2.0cm]{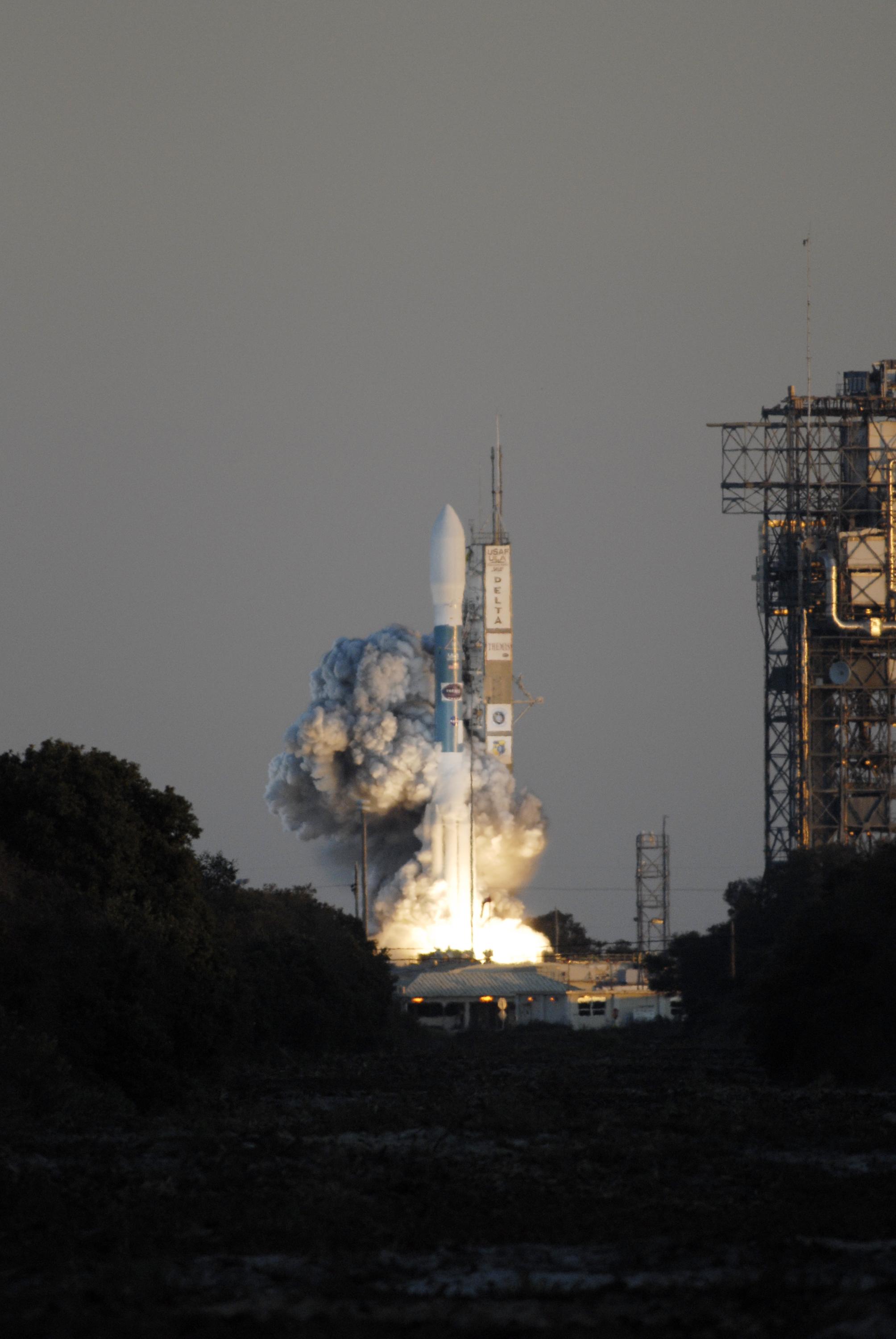}
    &\RaiseImage[height=2.0cm]{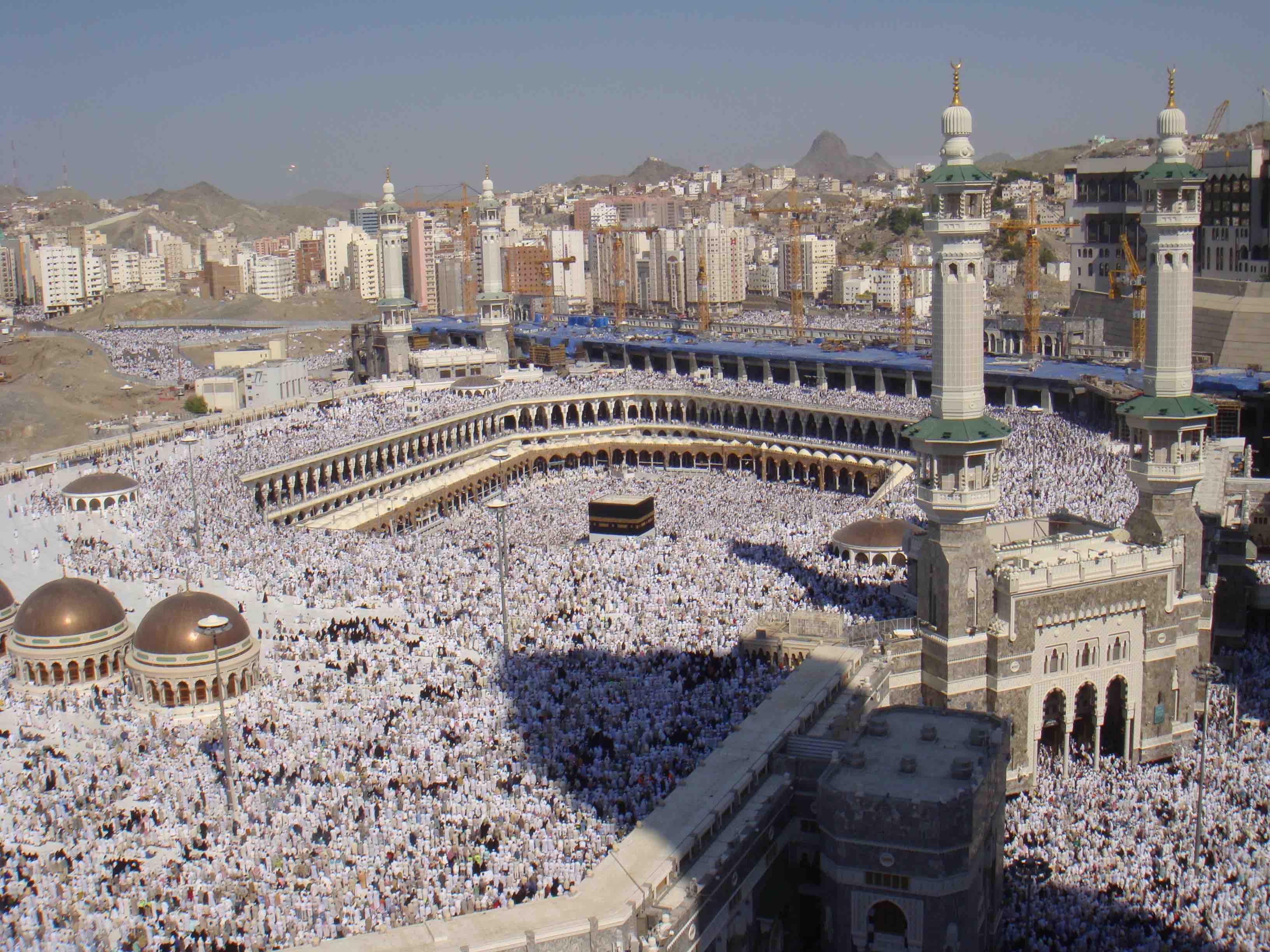}
    &\RaiseImage[height=2.0cm]{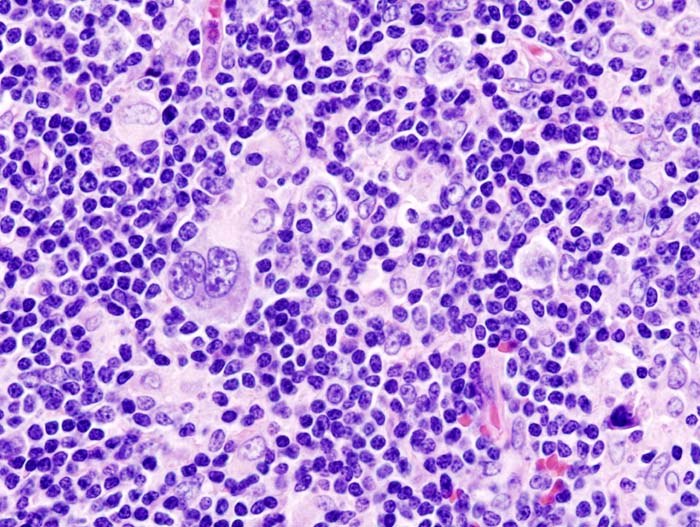}
\\
\textbf{Text}
    & Australians before the \ul{\textbf{competition}} started for the day at the bottom of the \ul{\textbf{hill}}
    & 
    spacecraft aboard lifts off at  \ul{\textbf{Cape Canaveral Air Force Station}} at 6:01 p.m
    & \ul{\textbf{pilgrims}} at Mecca's Grand Mosque in 2008
    & \ul{\textbf{Histopathologic image}} of Hodgkin's lymphoma.
\\ \midrule
\textbf{GT}
    & \makecell{Nor-Am Cup ; Copper \\ Mountain (Colorado)}
    & \makecell{Cape Canaveral\\Space Force Station}
    & \makecell{nil}
    & \makecell{Histopathology}
\\
\textbf{Pred (L+V)} 
    & \makecell{Competition ;\\The Hill ( Film )}
    & \makecell{Cape Canaveral\\Air Force Station}
    & \makecell{Pilgrims ( Plymouth \\ Colony )}
    & \makecell{nil}
\\ 
\bottomrule
\end{tabular}
}
\caption{Examples for (a) retrieval error, (b) misidentification, (c) over prediction, and (d) under prediction. }
\label{tab:error_examples}
\end{center}
\end{table*}

\begin{table*}[ht!]
\begin{center}
\scalebox{0.7}{
\begin{tabular}{cLVAM}
\toprule
    \textbf{ID}&\textbf{Error Type} & \textbf{Text} & \textbf{GT}& \textbf{Prediction}  \\ 
\midrule
(a)&Column name ambiguity
    & show the \ul{\textbf{name}} and the release year of the song by the youngest singer .
    & singer \# song name
    & singer \# name\\
\midrule
(b)&Lack of inference
    & which model be \ul{\textbf{lighter}} than 3500 but not build by the ' Ford Motor Company ' ?
    & cars data \# weight
    & model list \# model\\
\midrule
(c)&Prime key confusion
    & find the \ul{\textbf{id}} of the pet own by student whose last name be ` Smith ' .
    & has pet \# pet id
    & pets \# pet id\\
\midrule
(d)&Unknown strings 
    & ... list the car \ul{\textbf{makeid}} and make name .
    & car names \# make id
    & car names \# make\\
\bottomrule
\end{tabular}
}
\caption{Case study on schema linking errors. The hash symbol connects the table name and column name.}
\label{tab:error_examples_schema}
\end{center}
\end{table*}

\subsection{Error Analysis}
Figure \ref{fig:error_breakdown} shows the error breakdown on WikiDiverse. The errors are divided into four categories:  retrieval error where the target entity is not in the candidates; misidentification where the prediction does not match the ground truth entity; under predict where the model predicts ``nil'' and the ground truth entity is not ``nil''; over prediction where the ground truth entity is ``nil''.

Representative error examples are presented in Table \ref{tab:error_examples} for (a) retrieval error, (b) misidentification, (c)  over prediction, and (d) under prediction. {Error type (a) contributes to over half of the errors, emphasizing the need for a good retriever. It cannot be addressed by our model, because the ground truth entity is not in the set.
Example (b) is due to  candidate confusion, as Cape Canaveral Air Force Station is a previously used name for Cape Canaveral Space Force Station from 1974 to 1994 and from 2000 to 2020. Such errors indicate the necessity for a coreference system at inference time.
The over-prediction example as shown in (c) calls for a better discrimination strategy for plausible candidates. 
Example (d) is a challenging example, which asks future models to possess more profound prior knowledge. }

Table \ref{tab:error_examples_schema} further shows four types of errors for schema linking, name ambiguity, inference difficulty, prime key confusion, and unknown strings. 
(a) Name ambiguity is a common challenge for schema linking especially when the column names have overlapped tokens. (b) Sometimes the model fails to make inferences on subtle entity expressions. (c) Another common error type for schema linking is the confusion in prime keys, {as the prime key ``pet id'' is shared by multiple tables in the example}. (d) Another challenge is unknown strings or composite tokens since it is usually intractable to recover the original expression from those mentions.

\begin{table*}[ht!]
\begin{center}
\scalebox{0.66}{
\begin{tabular}{SAAAAM}
\toprule
\textbf{ID} & (a) & (b) & (c) & (d) & (e)
    \\ 
\textbf{Image}
    &\RaiseImage[height=2.0cm]{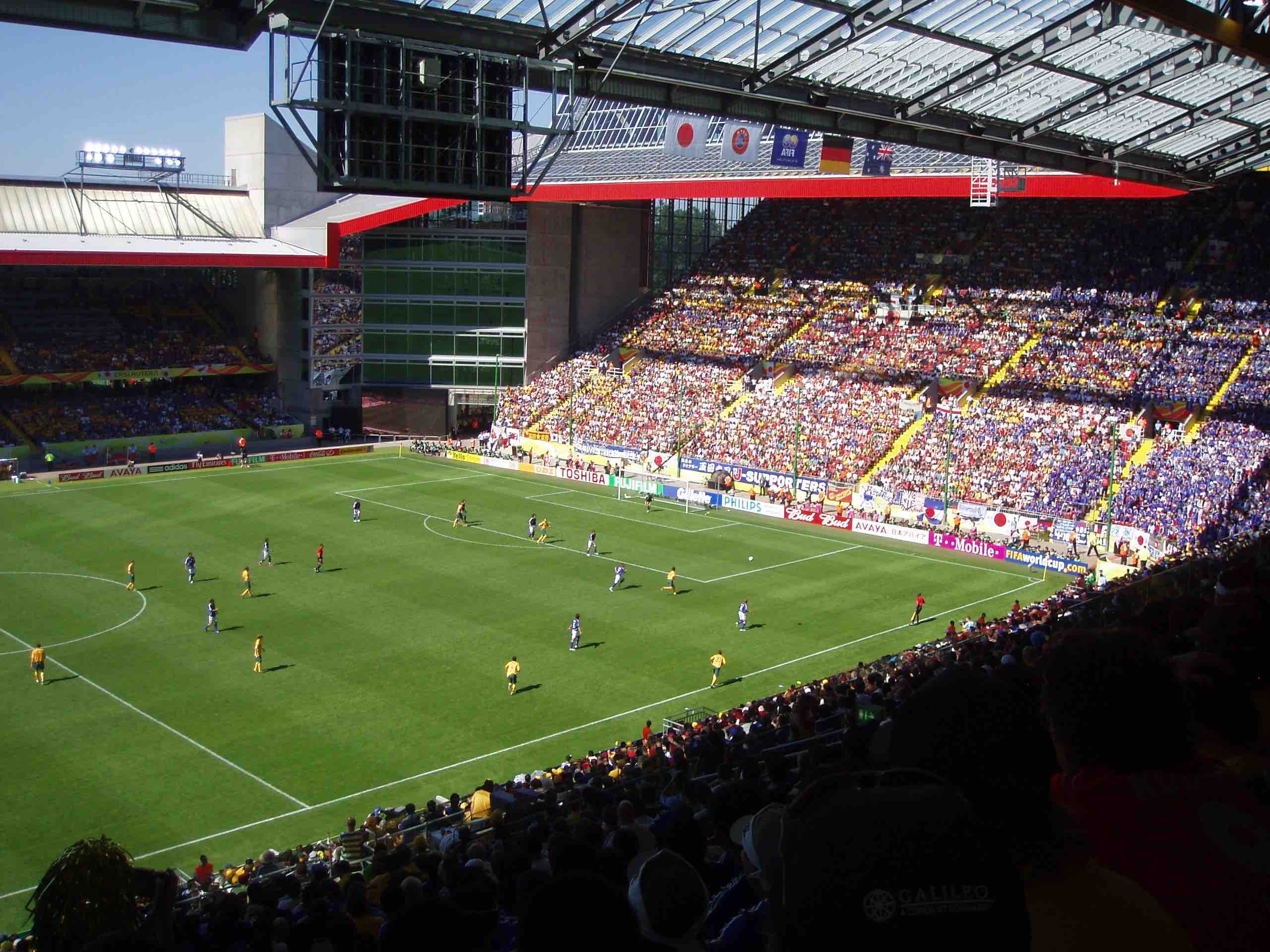}
    &\RaiseImage[height=2.0cm]{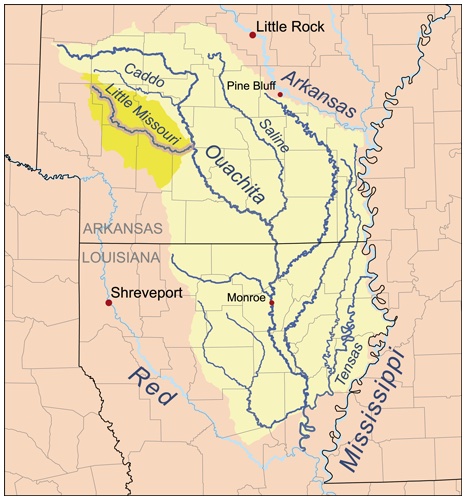}
    &\RaiseImage[height=2.0cm]{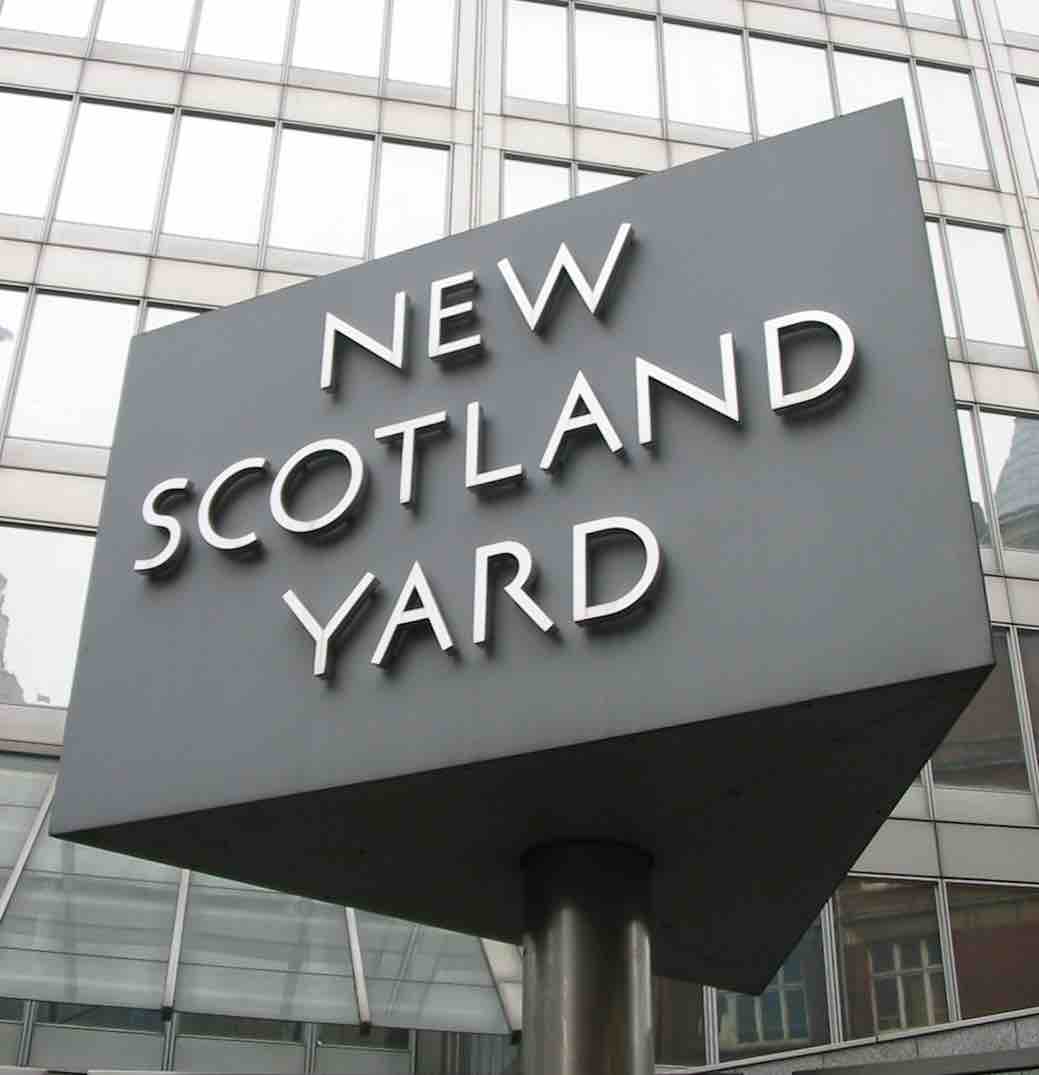}
    &\RaiseImage[height=2.0cm]{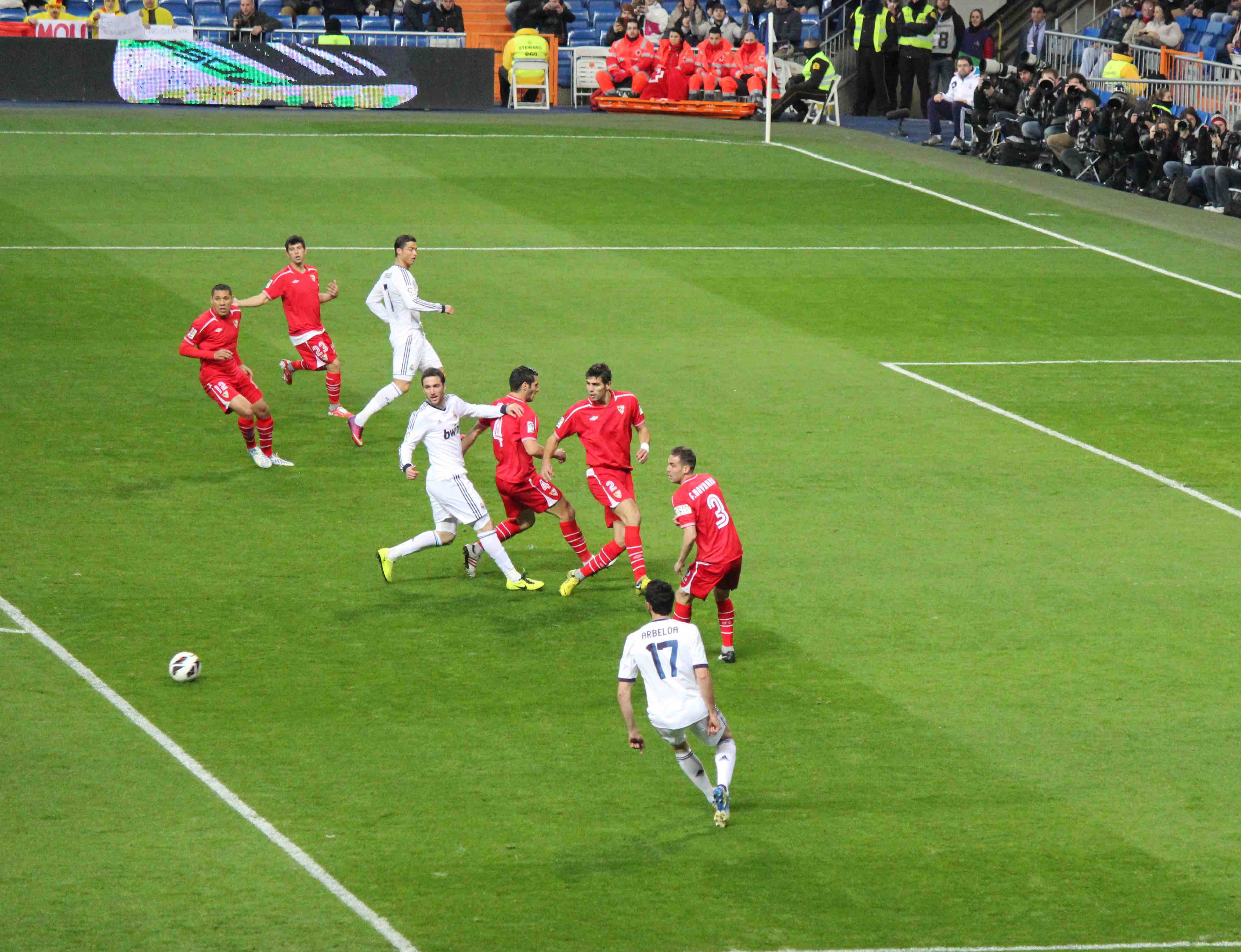}
    &\RaiseImage[height=2.0cm]{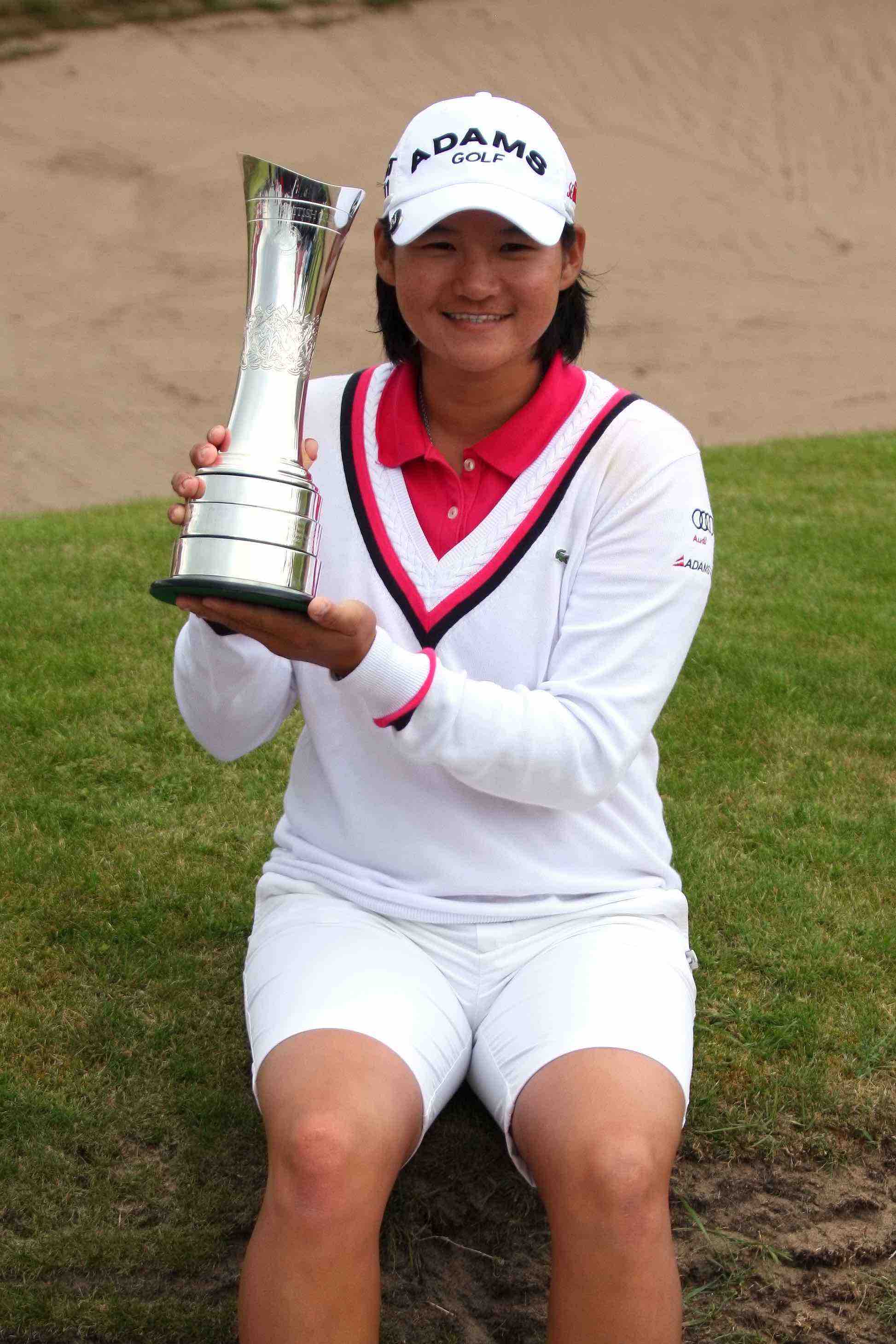}
\\
\textbf{Text}
    & \ul{\textbf{Australia}} and \ul{\textbf{Japan}} at the 2006 World Cup
    & ... around the \ul{\textbf{Little Missouri}} (highlighted) and Caddo rivers.
    & ... iconic sign at current Metropolitan Police \ul{\textbf{headquarters}}.
    & Real Madrid players including \ul{\textbf{Ronaldo}} go for the ball following a foul.
    & \ul{\textbf{Tseng}} with the trophy
\\ \midrule
\textbf{Pred(L+V) =GT}
    &Australia national soccer team ; Japan national football team
    & Little Missouri River (Arkansas)
    & Scotland Yard
    & Cristiano Ronaldo
    & Yani Tseng
\\
\textbf{Pred (L)} 
    &Australia ; Japan
    & Little Missouri River Bridge
    & Headquarters
    & Ronaldo ( Brazilian Footballer )
    & Sam Tseng
\\ 
\bottomrule
\end{tabular}
}
\caption{ Ablation on when the visual content is indispensable for entity disambiguation. 
}
\label{tab:pred_comp}
\end{center}
\end{table*}

\subsection{Ablation Study} 
\label{sec:ablation}
{We further discuss the indispensability of information within each modality through an ablation study on WikiDiverse. {Three evaluations are conducted under settings: (1) L+V where both text and image are leveraged; (2) L where only texts are used for prediction; (3) V where only images are used for prediction.}
The experimental results are reported in Table \ref{tab:wikidiverse}. It shows each model achieves the best performance with both text and image modalities (L+V).  While textual content provides the most inference clues, visual content provides complementary information. }
{Additionally, the proposed model \Model outperforms LXMERT with a single generic model trained for various  configurations, while three modality-specific models 
for each  configuration (L+V, L, V) are trained in LXMERT. This observation demonstrates the effectiveness of the proposed model for various modality configurations.}

Table \ref{tab:pred_comp} shows several misinformation examples where the model fails without visual information.
In (a), the image of a soccer stadium provides extra semantics when the model misses the semantic indicator from ``the 2006 World Cup.'' The image in example (b) is in Wiki-KB and has been used for pre-training. Its visibility during the pre-training makes the image a strong indicator of the target entity. The optical character ``NEW SCOTLAND YARD'' in example (c) is indispensable for the mention to be correctly identified. Without the optical features, inferring the mention of ``headquarters'' is challenging given the ambiguous text. Examples (d) and (e) emphasize the dependence on facial features for entity disambiguation. In the absence of the image in (d), it is impossible to disambiguate between ``Cristiano Ronaldo'' and ``Ronaldo ( Brazilian Footballer )'' as both players served Real Madrid. 

\begin{table}[t]
\centering
\small
\scalebox{0.9}{
\begin{tabular}{X|c|c|c}
\toprule
    \multirow{2}{*}{\textbf{Method}}  & \multicolumn{3}{c}{\textbf{F1}}\\
    \cmidrule(lr){2-4}
    & L+V & L & V\\
\midrule
    LXMERT \cite{wang-etal-2022-wikidiverse} & 71.07& 63.65& 40.16\\
\midrule

{\textbf{\Model-base}} &  \textbf{79.10}{\scriptsize$\pm$0.35}  & 76.79{\scriptsize$\pm$0.32} & \textbf{40.59}{\scriptsize$\pm$2.32}\\
{\textbf{\Model-large}}&  78.69{\scriptsize$\pm$0.33} & \textbf{77.05}{\scriptsize$\pm$0.29} & 37.65{\scriptsize$\pm$0.25}\\
\bottomrule
\end{tabular}}
\caption{Ablation study on WikiDiverse. 
}
\label{tab:wikidiverse}
\end{table}

\section{Related Works}
\label{sec:related_work}


\paragraph{Textual Entity Linking}

Early entity linking researches \cite{hoffart-etal-2011-robust, Daiber2013Improving} reply on probabilistic approaches, based on textual similarity and corpus occurrence. 
A more recent line of research is neural networks based retrieval-reranking approaches \cite{9288280, zhang2022entqa, mrini-etal-2022-ddr}, which first retrieve top candidates given the input text, and then score each candidate with semantic similarity or correlation.
End-to-end entity linking {models \cite{broscheit-2019-investigating, martins-etal-2019-joint, 9288280} approach this problem by directly detecting the entity mentions and linking them to their corresponding entities in the KB}.
For example, autoregressive entity linking models \cite{decao2021genre, de-cao-etal-2021-highly, petroni-etal-2021-kilt, mrini-etal-2022-ddr} formulate entity linking as a language generation problem using an encoder-decoder model. 

\paragraph{Textual-Visual Entity Linking}
The growing trend towards multimodality significantly advanced research in multimodal entity linking. 
Due to the difficulty in collecting and cleaning multimodal entity linking data, previous researchers limit their attention to a specific domain such as social media data \cite{twitter_mel1, twitter_mel2, moon-etal-2018-multimodal-snap, gan_2021_melbench} and news domain\cite{Zheng_2022_veld,wang-etal-2022-wikidiverse}, or a limited scope like person and organization recognition \cite{gan_2021_melbench, cui2021:whos-waldo}.
Previous work \cite{wang-etal-2022-wikidiverse} represents each entity with one image, which limits the visual expression of entities. We overcome this limitation by pre-training \Model with multiple images per entity to obtain diverse visual representations.

\paragraph{Tabular Schema Linking} Schema linking \cite{Guo2019TowardsCT, wang-etal-2020-ratsql} is an instance of entity linking in the context of linking to the relational database schema. Previous research shows that good schema linking \cite{liu-etal-2021-awakening, Katsakioris_2022_tabular_el_needs_attention, shi-etal-2020-squall, lei-etal-2020-slsql, chen-etal-2020-tale} can substantially improve downstream tasks such as Text-to-SQL parsing.  However, entity mentions in existing benchmarks such as Spider \cite{yu-etal-2018-spider} can almost exactly match the corresponding schema entities \cite{chen-etal-2020-tale}. 
Therefore, current Text-to-SQL semantic parsers normally address this problem with string-matching heuristics \cite{chen-etal-2020-tale} or embedding matching modules \cite{chen-etal-2020-tale, Wang2022ACM_proton, Guo2019TowardsCT, wang-etal-2020-ratsql}. 
However, due to the diversity and ambiguity in natural language mentions, such heuristics are hard to generalize to new domains \cite{chen-etal-2020-tale, Wang2022ACM_proton}.

\paragraph{Multimodal Models}
Multimodal models have attracted increasing attention in computer vision and natural language processing communities.
Recent transformer-based approaches \cite{vilt, Radford2021clip, flava} that leverage the attention between the visual and textual embeddings manifest the effectiveness of the attention mechanism. However, the proposed learning objectives are usually limited to predefined scopes, such as text-image matching or alignment \cite{xu-etal-2021-layoutlmv2, Radford2021clip,Biten2022LaTrLT, ho2022yoro,li2022blip, Yang2022tcl, huang2023nlip}, semantic segmentation, object detection, classification \cite{xu2022groupvit, guo2022visual, assran2022ViTMSN},  and masked language modeling \cite{li-etal-2020-bert-vision, xclip, tong2022videomae, DocFormer}. Instead, we proposed a generic generative model that is open to diverse downstream tasks.  
Additionally, \Modelsp differs from previous generative multimodal models\cite{li2023trocr, Wang2021speechencoder} in that \Modelsp can process and comprehend information from heterogeneous instead of a single source;
\Modelsp differs from VL-T5 \cite{cho2021vlt5} and others \cite{Wang2022UnifyingAT, bao2022beit, beit3} in that \Model enables thorough encoding for each modality, instead of discretizing visual content.

\section{Conclusion}
\label{sec:conclusion}
In this paper, a novel \task problem is formulated, which links the entity mention within heterogeneous information to a defined KB.
A generic \task dataset is built covering diverse EL tasks. 
We propose a unified generative model for \task, \Model. 
Comprehensive experiments are conducted over the \task dataset. Experimental results show that the proposed \Modelsp outperforms state-of-the-art models on almost each individual EL task.

\paragraph{Broader Impact}
In contrast to previous work \cite{pan2022contraqa, ChatGPT} that only leverage textual content, the proposed model has the potential to deal with misinformation (text only as a data source might be prone to misinformation or fake context/information).
This research will lead to a clearer understanding of misinformation issues and encourage better leverage of multimodal information. 


\section*{Limitations}

\Model establishes a compelling starting point for \task research. In spite of this, the proposed approach has several shortcomings. First, \Model currently generates entity name within the entity candidate set, however, we saw how retrieval errors limit entity linking performance. Thus, how to work collectively with the retrieval system to diminish errors takes appropriate action. Second, how to handle large tables still remains under-explored. It is infeasible to represent a huge database with the table flattening technique. In practice, it is possible to filter out less likely candidates to compress the search space, but a more promising approach is to represent the table more efficiently.

\Model also enables studies on more diverse-modal tasks. New tasks can be easily framed based on the proposed architecture, such as visual question answering, grounded generation, and diverse-modal commonsense reasoning. We believe that with more follow-up work on diverse tasks, this approach will turn out to be a more comprehensive generative diverse-modal framework.

\bibliography{ref}
\bibliographystyle{acl_natbib}

\appendix
\clearpage


\section{\task Selection}
The \task benchmark's datasets were carefully chosen after conducting extensive research on publicly available datasets. The \task benchmark includes five datasets: GERBIL, WikiDiverse, MELBench, Squall, and SLSQL. Each of these datasets is necessary and represents the best option for a comprehensive evaluation for Diverse-Modal Entity Linking. 
The selected datasets best align with the \task problem, among more than 20 datasets we looked into. The reasons why other datasets are excluded in the benchmark are: i) most dataset collected from social media cannot be reproduced because some of the data are no longer accessible. This category includes Twitter-MEL \cite{twitter_mel1}, SnapCaptionsKB\cite{moon-etal-2018-multimodal-snap}; ii) Dataset is not publicly available \cite{Zheng_2022_veld}; iii) Annotated dataset for schema linking is limited. Existing work that investigates entities in tables are Text-to-SQL datasets. However, annotations for schema linking are not available, such as Spider \cite{yu-etal-2018-spider}, Spider-Syn \cite{gan-etal-2021-towards}, Spider-DK \cite{gan-etal-2021-exploring}, WikiSQL\cite{zhongSeq2SQL2017}, ATIS \cite{price-1990-evaluation}, Freebase917 \cite{cai-yates-2013-large},  and WikiTableQuestions \cite{pasupat-liang-2015-compositional}. Furthremore, there is no overlap between the datasets. For tabular datasets, Squall is built upon WikiTableQuestions, while SLSQL is based on the Spider text-to-SQL dataset. Lastly, GENRE is widely recognized as the standard dataset for the task of textual entity linking.

\section{Implementation Details}
\label{sec:Implementation}
For every dataset in \task dataset, the fine-tuning procedure runs for 5 epochs with a batch size of 16. For both pre-training and fine-tuning, the learning rate is $3\times10^{-5}$, with a linear scheduler with 0.1 warmup ratio. Fine-tuning takes 5 hours for WikiDiverse, 2 for MELBench-Wiki, 1 for Squall, and 2 for SLSQL.

We run each setting 5 times and report the mean and variance unless stated otherwise (except the zero-shot setting when evaluated with the pre-trained checkpoint, since there is no randomness with the pre-trained checkpoint ). One MELBench, since the dataset split is not given along with the released MELBench-Wikidata data, we randomly split the dataset according to their split statistics and repeat the experiment 5 times to get average evaluation metrics. On Squall, we reported a 5-fold cross-validation result following the released split. For Squall and SLSQL, all the hyperparameters are tuned on the training set since there is no test set.

Additionally, Wikipedia images are collected through hyperlinks shared by \citet{wang-etal-2022-wikidiverse} at \url{https://github.com/wangxw5/wikidiverse}. 
{Annotations on SLSQL are adapted from Spider, excluding train\_others.json that are from Restaurants, GeoQuery, Scholar, Academic, IMDB, and Yelp prepared by \citep{P18-1033_yelp}.} For GERBIL benchmark results, we report average F1 scores on six test sets, including Aida-test, MSNBC-test AQUAINT-test, ACE2004-test, WNED-CWEB-test, and WNED-WIKI-test following \citet{decao2021genre}.

\section{Detailed Experimental Results}
\label{sec:appendix:experiments}
Experimental results for each individual dataset are shown in this section. Specifically, Table \ref{tab:melbench} shows results for MELBench, Table \ref{tab:squall} shows the experimental result for Squall, and Table \ref{tab:slsql} shows the experimental result for SLSQL.


\section{Domain Adaption Results}
We investigate zero-shot performance in unseen domains on the WikiDiverse dataset.  Specifically, we choose the five domains as seen domains, including politic, crime, sports, entertainment, and technology, and the rest five domains as the unseen domains, including disaster, health, economy, weather, and education. Training data includes instances from the seen domain, and the instances from the unseen domains are randomly split into validation and test set. Note that the data used for this experiment is from the WikiDiverse training and validation set, the data in the test set are excluded.

The domain adaption result on WikiDiverse is shown in Table \ref{tab:zero_wikidiverse}. These experiments facilitate studying knowledge transfer between seen and unseen domains. The experiment results show that (a) pre-training is indispensable for new domains as it provides profound prior knowledge for the MEL task in general; (b) knowledge learned from seen domains can indeed transfer to the unseen domain as the average F1 score improves by 3.61 percentage points.



 
\begin{table}[th]
\centering
\small{
\begin{tabular}{N|c|c}
\toprule
    \multirow{2}{*}{\textbf{Method}} &  \multicolumn{2}{c}{\textbf{F1}}\\ 
\cmidrule(lr){2-3}
    &Top-1&Top-10\\
\midrule
    MELBench \cite{gan_2021_melbench} & 40.5 & 69.6\\
\midrule
\textbf{\Model-base}  & 	68.01{\scriptsize$\pm$0.75} & 73.31{\scriptsize$\pm$0.77}\\
\textbf{\Model-large} & \textbf{72.41}{\scriptsize$\pm$0.65} &  \textbf{76.34}{\scriptsize$\pm$0.78}              
\\
\bottomrule
\end{tabular}
}
\caption{Results of entity linking on MELBench}
\label{tab:melbench}
\end{table}

\begin{table}[th]
\centering
\small
\begin{tabular}{ll|S}
\toprule
    \textbf{Method}& &  \textbf{F1}\\ 
\midrule  
    GENRE && 75.92{\scriptsize$\pm$4.29}\\
    GENRE+&& 82.10{\scriptsize$\pm$2.41}\\
\midrule
    \multirow{2}{*}{\textbf{\Model-base}}& Zero-shot&47.52{\scriptsize$\pm$1.06}  \\
      & Finetuned&\textbf{89.69}{\scriptsize$\pm$0.77} 
      \\
    \multirow{2}{*}{\textbf{\Model-large}} & Zero-shot& 49.14{\scriptsize$\pm$1.26} \\
    & Finetuned&89.12{\scriptsize$\pm$1.03}\\
\bottomrule
\end{tabular}
\caption{Results of schema linking on Squall. GENRE+ denotes augment table representations to the input text as described in Section \ref{sec:input_processor}}
\label{tab:squall}
\end{table}

\begin{table}[th]
\centering
\small
\begin{tabular}{ll|S}
\toprule
    \textbf{Method}& &  \textbf{F1}\\ 
\midrule  
    GENRE & & 70.41\\
    GENRE+& & 82.80 \\
\midrule
    \multirow{2}{*}{\textbf{\Model-base}}& Zero-shot& 30.92  \\ 
      & Finetuned& 81.48{\scriptsize$\pm$1.06}
      \\
    \multirow{2}{*}{\textbf{\Model-large}} & Zero-shot& 28.44 \\
    & Finetuned&\textbf{84.43}{\scriptsize$\pm$0.92}\\
\bottomrule
\end{tabular}
\caption{Results of schema linking on SLSQL}
\label{tab:slsql}
\end{table}

\begin{table}[th]
\centering
\small
\begin{tabular}{c|ccc}
\toprule
   \multirow{2}{*}{\textbf{Domain}} &  \multicolumn{3}{c}{\textbf{F1}}\\ 
   & FT w/o PT&    PT  & FT with PT\\
\midrule
    Health      & 40.37     &  76.19    &  82.48\\
    Weather     & 39.80     &  79.79    &  78.84\\
    Economy     & 46.67     &  78.95    &  84.17\\
    Disaster    & 34.67     &  78.81    &  81.04\\
    Education   & 40.23     &  73.49    &  81.48\\
\midrule
    Overall     & 39.17     &  78.05    &  81.66\\
\bottomrule
\end{tabular}
\caption{Evaluation on unseen domains in WikiDiverse with \Model-base. FT and PT stand for fine-tuning on seen domains and pre-training.}
\label{tab:zero_wikidiverse}
\end{table}


\end{document}